%% file: main_arxiv.tex
\PassOptionsToPackage{dvipsnames}{xcolor}
\documentclass[11pt,letterpaper]{style}

\usepackage[numbers]{natbib}
\usepackage{graphicx}
\usepackage{booktabs}
\usepackage{amsmath,amsfonts,amssymb}
\usepackage{cleveref}
\usepackage{subcaption}
\usepackage{wrapfig}
\usepackage{multirow}
\usepackage{colortbl}
\usepackage{listings}
\usepackage{xparse}
\usepackage{fontawesome5}
\usepackage{bxcoloremoji}
\usepackage{float}
\usepackage{placeins}
\usepackage{threeparttable}

\graphicspath{{./}{fig/}{figures/}{plot/}{pdf/}{table/}}
\usepackage{amsthm}
\usepackage{tcolorbox}
\usepackage{svg}
\tcbuselibrary{skins,breakable}
\tcbuselibrary{listingsutf8}
\usepackage{titletoc}

\usepackage{setspace}
\usepackage{pifont}
\usepackage{mathtools}
\usepackage{enumitem}
\usepackage{arydshln}
\usepackage{bbm}
\usepackage{lineno}
\usepackage{makecell}
\usepackage{adjustbox}
\usepackage{algorithm}
\usepackage{algorithmic}
\usepackage{caption} 
\usepackage{wrapfig}
\usepackage{hyperref}
\usepackage{bbding}
\usepackage{xspace}

\newcommand{\cmark}{\CheckmarkBold}

\newcommand{\method}{LangMARL\xspace}

\hypersetup{colorlinks=true,linkcolor=red,urlcolor=blue,citecolor={blue}}

\newcommand{\hua}[1]{\textcolor{red}{[Hua-- #1]}}

\definecolor{mygray}{gray}{0.9}
\definecolor{syncol}{RGB}{243,246,249}
\definecolor{wildcol}{RGB}{215,240,235}
\definecolor{drop1}{RGB}{180,225,220}
\definecolor{drop2}{RGB}{150,210,200}
\definecolor{drop3}{RGB}{120,195,185}
\definecolor{drop4}{RGB}{95,180,170}
\definecolor{drop5}{RGB}{65,160,150}
\definecolor{lightblue}{RGB}{26,82,249}

\definecolor{myblue1}{HTML}{0171DC}
\definecolor{myblue2}{HTML}{013978}

\NewDocumentEnvironment{minted}{O{} m +b}{%
}{}

\newcommand{\equal}{\textsuperscript{*}}     

\renewcommand\Authfont{\centering\normalfont\bfseries\fontsize{11}{15}\selectfont}
\renewcommand\Affilfont{\centering\normalfont\fontsize{10}{15}\selectfont}

\title{LangMARL: Natural Language Multi-Agent Reinforcement Learning}
\runningtitle{LangMARL: Natural Language Multi-Agent Reinforcement Learning}

\author{%
    {\Authfont
    \textbf{Huaiyuan Yao}\textsuperscript{*1} \quad
    \textbf{Longchao Da}\textsuperscript{*1} \\ 
    \textbf{Xiaoou Liu}\textsuperscript{1} \quad
    \textbf{Charles Fleming}\textsuperscript{2} \quad
    \textbf{Tianlong Chen} \textsuperscript{3} \quad
    \textbf{Hua Wei
 }\textsuperscript{1}
    }\\
    {\Affilfont
    \textsuperscript{1}  Arizona State University \quad \\
     \texttt{\{huaiyuan, longchao, xiaoouli, hua.wei\}@asu.edu} \\
   \textsuperscript{2}  Cisco Research \quad 
     \texttt{chflemin@cisco.com} \\
    \textsuperscript{3}  University of North Carolina at Chapel Hill \quad
     \texttt{tianlong@cs.unc.edu} \\
    \textsuperscript{*} Equal Contribution
    }
}

\begin{document}
\input{sec_arxiv/0_abstract}
\newcommand{\TitleLinks}{%
\centering
    \vspace{8pt}
}
\maketitle

\input{sec_arxiv/1_intro}
\input{sec_arxiv/2_related}

\input{sec_arxiv/3_preliminary}
\input{sec_arxiv/4_method}

\input{sec_arxiv/5_exp}

\input{sec_arxiv/6_conclusion}

\bibliographystyle{unsrtnat}  
\bibliography{ref}

\appendix
\input{sec_arxiv/appendix}

\end{document}

%% file: sec_arxiv/0_abstract.tex
\begin{abstract}
Large language model (LLM) agents struggle to autonomously evolve coordination strategies in dynamic environments, largely because coarse global outcomes obscure the causal signals needed for local policy refinement. We identify this bottleneck as a multi-agent credit assignment problem, which has long been studied in classical multi-agent reinforcement learning (MARL) but remains underaddressed in LLM-based systems. Building on this observation, we propose LangMARL, a framework that brings credit assignment and policy gradient evolution from cooperative MARL into the language space. LangMARL introduces agent-level language credit assignment, pioneers gradient evolution in language space for policy improvement, and summarizes task-relevant causal relations from replayed trajectories to provide dense feedback and improve convergence under sparse rewards. Extensive experiments across diverse cooperative multi-agent tasks demonstrate improved sample efficiency, interpretability, and strong generalization. Our code and document are available at \url{https://langmarl-tutorial.readthedocs.io/}
\end{abstract}

%% file: sec_arxiv/1_intro.tex

\section{Introduction}

\begin{wrapfigure}{r}{0.5\textwidth} 
    \centering
    \includegraphics[width=0.48\textwidth]{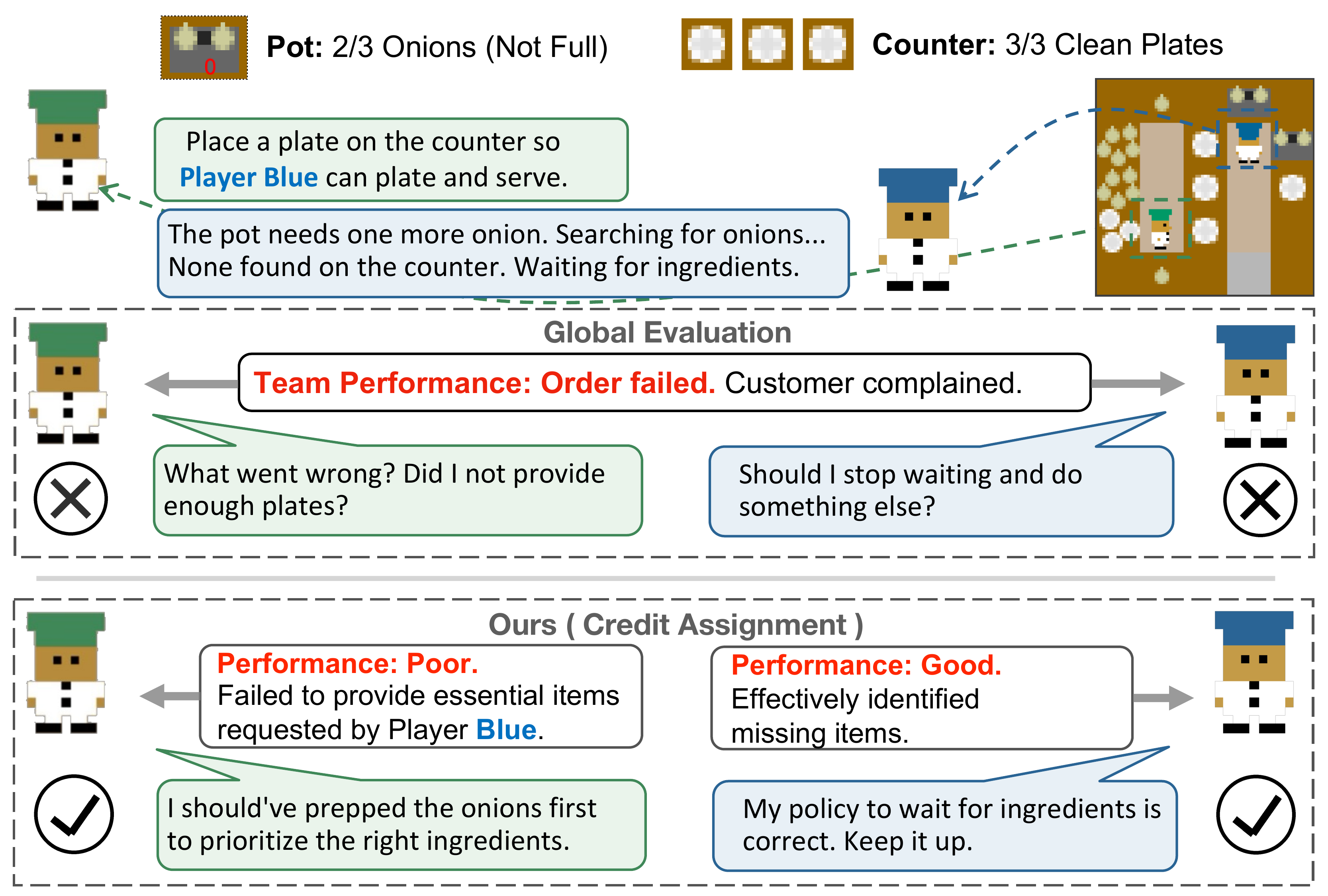} 
    \caption{\textbf{Challenges in multi-agent credit assignment.} Global evaluation fails to pinpoint individual contributions, leading to ambiguous reflections. LangMARL addresses this by decomposing team performance into agent-specific credits.}
    \label{fig:challenge}
\end{wrapfigure}

LLMs have advanced multi-agent systems (MAS) by enabling effective collaboration on diverse complex tasks~\cite{li2024survey, zhao2025autonomousmultimodalllmagents}. However, deploying these systems in dynamic environments remains challenging due to a lack of principled adaptive mechanisms. Most current MAS architectures rely on manually engineered prompts and static configurations, which prevent agents from evolving their strategies as task distributions change~\cite{yao2025comalcollaborativemultiagentlarge, guo2024largelanguagemodelbased}. A deeper obstacle is that global outcomes often provide insufficient guidance for local policy refinement, as shown in Fig.~\ref{fig:challenge}. Without attributing team performance to individual agents, existing systems resort to coarse-grained optimization such as global reflections or monolithic prompt rewrites~\cite{bo2024reflective, ma2025agentic}, leading to inefficient learning or coordination collapse. This bottleneck is fundamentally a \textit{credit assignment} problem: \textit{how to derive effective learning signals for each agent?} While well-studied in MARL~\cite{foerster2018coma}, it remains largely overlooked in LLM-based multi-agent systems~\cite{li2025multi}.

\begin{figure*}[t]
  \centering
  \includegraphics[width=1.0\textwidth]{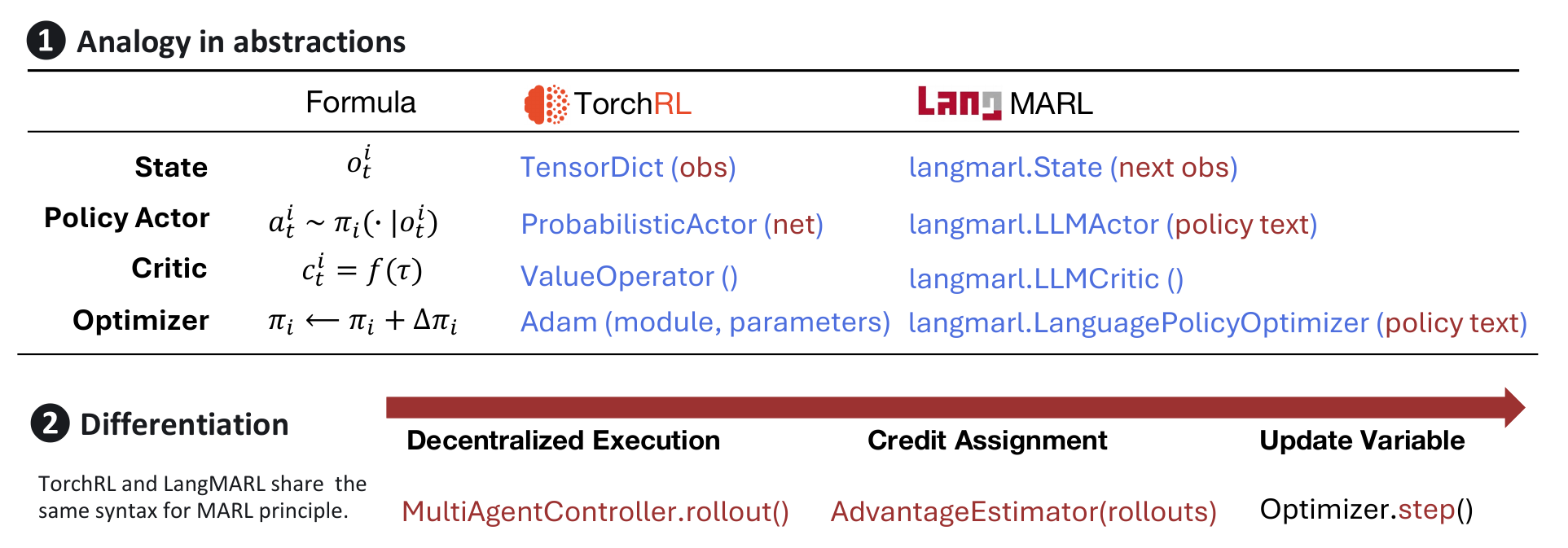}
  \caption{\textbf{An Easy-to-Use Toolkit for LangMARL.} LangMARL mirrors the syntax and abstractions of classical MARL libraries (e.g., TorchRL), redefining core components in natural language space, making LLM-based multi-agent optimization as straightforward to implement as standard deep RL pipelines.}
  \label{fig:theoretical_mapping}
\end{figure*}


These gaps can be addressed by incorporating mature wisdom from the MARL field, which offers principled solutions to exactly these challenges. MARL formalizes multi-agent learning as an experience-driven optimization process~\cite{Tan1997MultiAgentRL}, and the `\textit{centralized training decentralized execution (CTDE)}' paradigm~\cite{lowe2017multi} enables agents to leverage global information during training while maintaining decentralized autonomy at execution. Crucially, MARL has developed mature credit assignment techniques such as counterfactual baselines~\cite{foerster2018coma} and value decomposition~\cite{sunehag2017value} that precisely attribute global outcomes to individual agents.

However, directly applying MARL techniques to LLM-based agents is non-trivial, as LLMs operate in discrete language spaces as black-box models, creating a fundamental mismatch with conventional numeric learning signals. To tackle this challenge, we introduce \textbf{LangMARL}, a framework that draws on the core principles of MARL and credit assignment to design an automatic optimization system for multi-agent LLMs. Concretely, LangMARL consists of the following components: \textit{(1) Language Policy Actors}, where each agent maintains a natural language policy and selects actions conditioned on textual observations; \textit{(2) Centralized Language Critic}, which observes complete episodic trajectories and performs causal credit assignment in natural language, attributing team outcomes to individual agents; \textit{(3) Language Policy Gradient Estimator}, which converts agent-specific language credits into language-form policy update directions, serving as an analogue of policy gradients; and \textit{(4) Language Policy Optimizer}, which aggregates language gradients and applies semantic policy updates through LLM-based optimization operators. By grounding these MARL-inspired components, LangMARL places credit assignment at the core of multi-agent LLM optimization, where the Centralized Language Critic attributes global outcomes to individual agents and drives all downstream policy updates, enabling principled autonomous evolution of LLM-based multi-agent systems. 
Our contributions are summarized as follows.
\begin{itemize}
    \item We identify credit assignment as the central bottleneck for autonomous optimization in multi-agent LLM systems, and introduce a paradigm that draws on core MARL principles to enable LLM-based agents to autonomously learn and evolve coordination strategies. 
    \item We propose LangMARL, a general framework that instantiates MARL-inspired automatic optimization for multi-agent LLMs, featuring language-parameterized policies, centralized credit assignment, and a credit-driven language policy optimizer. We modulate the framework into an easy-to-implement toolkit like TorchRL, as depicted in Fig.~\ref{fig:theoretical_mapping}.
    \item Extensive experiments across reasoning, QA, coding, and game tasks demonstrate the effectiveness and scalability of LangMARL, with strong performance maintained across varying numbers of agents.
\end{itemize}

\begin{table*}[t]
\setlength{\tabcolsep}{12pt}
\renewcommand{\arraystretch}{0.95}
\centering
\caption{Comparison across prompting and self-evolving paradigms. LangMARL provides the most comprehensive framework by unifying self-evolution, centralized critique, and explicit credit assignment.}
\label{tab:self-evolving-compare}
\resizebox{0.98\textwidth}{!}{%
\begin{tabular}{llcccc}
\toprule
\textbf{Paradigm} & \textbf{Method} &
\textbf{Self-Evolving} &
\textbf{Central Critic} &
\textbf{Text Gradient} &
\textbf{Credit Assignment} \\
\midrule

\multirow{2}{*}{Static Prompting}
& CoT~\cite{wei2022chain}& \ding{56} & \ding{56} & \ding{56} & \ding{56} \\
& Agents~\cite{zhou2023agents}             & \ding{56} & \ding{56} & \ding{56} & \ding{56} \\
\midrule

\multirow{4}{*}{Single-Agent Self-Evolving}
& Auto Prompt Engineer (AutoPE)~\cite{zhou2023large}   & \cmark & \cmark & \ding{56} & \ding{56} \\
& DSPy~\cite{khattab2023dspy}               & \cmark & \ding{56} & \ding{56} & \ding{56} \\
& Reflexion~\cite{shinn2023reflexion}          & \cmark & \cmark & \ding{56} & \ding{56} \\
& TextGrad~\cite{yuksekgonul2024textgrad}           & \cmark & \cmark & \cmark & \ding{56} \\
\midrule

\multirow{3}{*}{Multi-Agent Self-Evolving}
& Agent Neural Network$^\dagger$~\cite{ma2025agentic}         & \cmark & \cmark & \cmark & \ding{56} \\
& Symbolic~\cite{ou2025symbolic}           & \cmark & \cmark & \cmark & \ding{56} \\
& \textbf{LangMARL (ours)}  & \textcolor{red}{\cmark} & \textcolor{red}{\cmark} & \textcolor{red}{\cmark} & \textcolor{red}{\cmark} \\
\bottomrule
\end{tabular}
}
\begin{flushleft}
\small $^\dagger$ Code is not publicly available at the time of writing.
\end{flushleft}
\vspace{-5mm}
\end{table*}


%% file: sec_arxiv/2_related.tex
\section{Related Work}

\subsection{Multi-agent LLMs}

LLM-based multi-agent systems have demonstrated strong collaborative capabilities across diverse domains \cite{wu2024autogen, yao2025instructional}, and recent work has begun to evaluate and survey LLMs' abilities in multi-agent coordination~\cite{tran2025multiagent_survey, yao2025instructionalagentsllmagents, zhao2025autonomousmultimodalllmagents}. However, most existing systems rely on static prompting approaches such as chain-of-thought reasoning \cite{wei2022chain} and manually designed agent pipelines, which depend heavily on human engineering and lack adaptive mechanisms. To reduce this reliance, automatic prompt engineering methods~\cite{yang2023large} have been integrated into agent systems, yet they optimize each agent independently without considering cooperative dynamics. And more multi-agent self-evolving frameworks~\cite{ou2025symbolic, zhao2025sirius, hua2025socianablatextualgradientmeets, hong2023metagpt} enable autonomous improvement by optimizing prompts, tools, and pipelines via language gradients.

\subsection{Prompt Optimization}
Prompt optimization aims to automatically discover or refine the textual instructions given to LLMs, reducing reliance on manual prompt engineering. 
Early work, such as Automatic Prompt Engineer \cite{zhou2023large}, shows that LLMs can generate instructions that rival those from human-designed prompts. DSPy \cite{khattab2023dspy} further abstracts LLM pipelines as declarative, modular programs with compiler-based prompt and demonstration optimization. A complementary direction explores learning through verbal feedback: Reflexion \cite{shinn2023reflexion} reinforces language agents via linguistic self-reflection stored in episodic memory, TextGrad \cite{yuksekgonul2024textgrad} backpropagates natural language feedback as textual gradients through computation graphs, and agent symbolic learning \cite{ou2025symbolic} extends language-based gradient descent to optimize entire agent pipelines. While these methods advance single-agent self-improvement, they do not address the multi-agent setting where credit must be disentangled among agents. LangMARL extends these self-evolving paradigms to multi-agent systems by introducing credit assignment, as shown in Table~\ref{tab:self-evolving-compare} and Appendix~\ref{app:paradigm}.

%% file: sec_arxiv/3_preliminary.tex
\section{Preliminaries}

\subsection{Cooperative MARL}
\label{sec:prelim:marl}

We consider a cooperative MARL setting, where $N$ agents interact with a shared environment to maximize a common team objective. The environment is modeled as a decentralized partially observable Markov decision process (Dec-POMDP), defined by the tuple
$
\mathcal{M} = \langle S, \{A_i\}_{i=1}^N, \{O_i\}_{i=1}^N, P, R, \gamma \rangle.
$
Here, $S$ denotes the global state space, $A_i$ and $O_i$ denote the action and observation spaces of agent $i$, respectively, $P$ is the state transition function, $R$ is a shared team reward function, and $\gamma \in [0,1)$ is the discount factor.
At each timestep $t$, the environment transitions according to
$
s_{t+1} \sim P(s_{t+1} \mid s_t, a_t^1, \ldots, a_t^N),
$
where $a_t^i \in A_i$ is the action taken by agent $i$. All agents receive a common reward
$
r_t = R(s_t, a_t^1, \ldots, a_t^N).
$ Each agent maintains a decentralized policy $\pi_i(a_i \mid o_i)$ that maps observations to actions. The joint policy is denoted by $\pi = (\pi_1, \ldots, \pi_N)$. The learning objective is to maximize the expected discounted return: $  J(\pi) = \mathbb{E}_{\pi}\left[ \sum_{t=0}^{T} \gamma^t r_t \right] $.

\subsection{Centralized Training with Decentralized Execution}
\label{sec:prelim:ctde}

A standard and effective paradigm for cooperative MARL is centralized training with decentralized execution (CTDE), which decouples training-time information access from execution-time constraints.
At execution time, each agent acts solely based on its local observation: $ a_t^i \sim \pi_i(\cdot \mid o_t^i) $, where $i=1,\ldots,N$. During training, a centralized critic is permitted to condition on global information,
including the full state $s_t$ and the joint action $a_t=(a_t^1,\ldots,a_t^N)$.
A typical realization is an action-value function: $ Q^\pi(s_t, a_t) =\mathbb{E}_{\pi}\!\left[ \sum_{k=t}^{T} \gamma^{k-t} r_k \,\middle|\, s_t, a_t \right]$, it evaluates the expected team return under the joint policy.

\subsection{Credit Assignment in Cooperative MARL}
\label{sec:prelim:credit}

A central challenge in cooperative MARL is \emph{credit assignment}: how to derive agent-specific learning signals from a shared team reward.
At each timestep $t$, all agents observe the same scalar reward
$
r_t = R(s_t, a_t),
$
where the joint action $a_t = (a_t^1,\ldots,a_t^N)$ collectively determines the team outcome.
However, effective policy optimization requires feedback that reflects each agent's individual contribution to this joint result. Rather than assuming an explicit per-agent reward decomposition, credit assignment is more generally formulated as the construction of
agent-specific learning signals:
\begin{equation}
    r_t \;\longrightarrow\; \{c_t^1, c_t^2, \ldots, c_t^N\}
\end{equation}
where $c_t^i$ denotes the signal used to update agent $i$'s policy.
Under an actor--critic formulation, the policy gradient for agent $i$ can be written as:
\begin{equation}
    \nabla_{\theta_i} J(\pi)
=
\mathbb{E}_{\pi}\!\left[
\sum_{t=0}^{T}
\nabla_{\theta_i}\log \pi_i(a_t^i \mid o_t^i)\, c_t^i
\right]
\end{equation} Making the design of $c_t^i$ central to how team-level performance is attributed to individual agents.

%% file: sec_arxiv/4_method.tex
\begin{figure*}[t]
  \centering
  \includegraphics[width=1.0\textwidth]{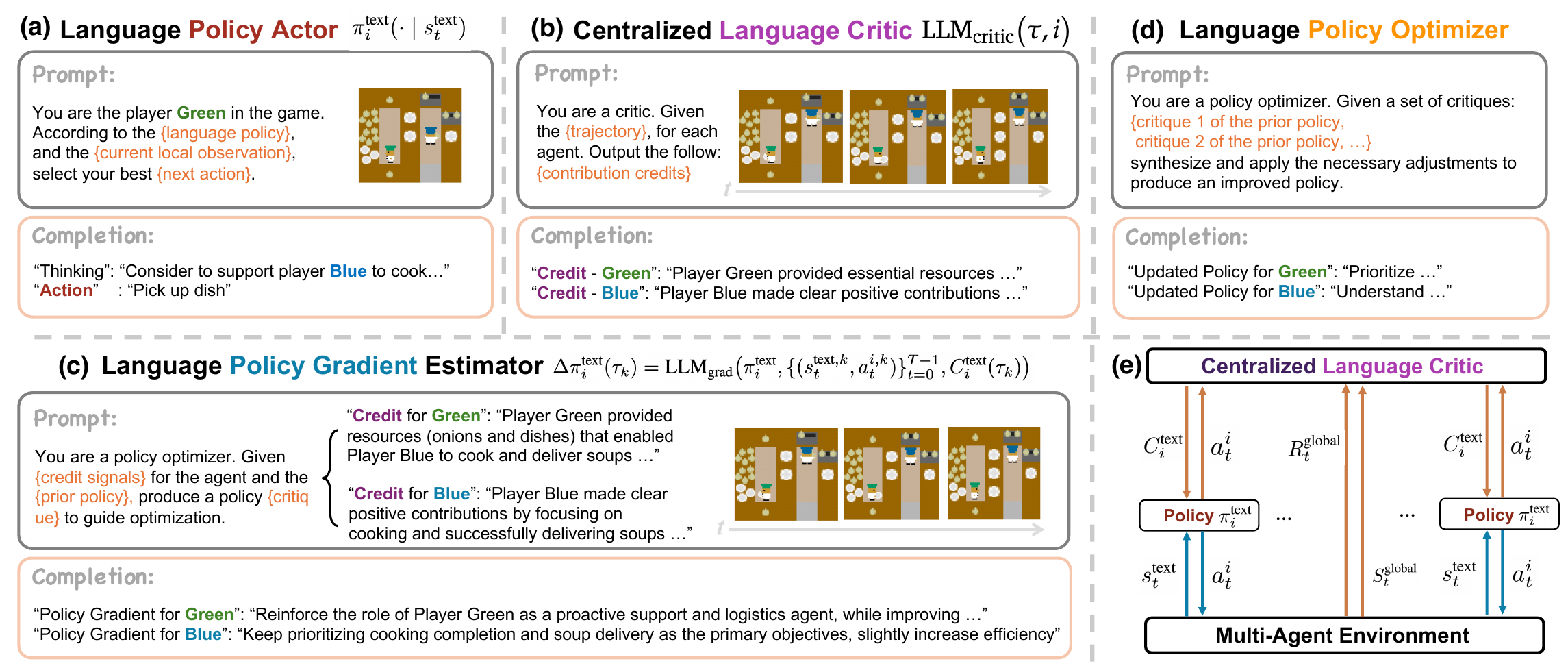}
  \caption{\textbf{The \method System Pipeline.} The framework follows a CTDE paradigm: (i) \textit{Language Policy Actors} execute decentralized actions, (ii) a \textit{Centralized Language Critic} assigns trajectory-level causal credits, and (iii) the \textit{Language Policy Optimizer} updates policies in natural language.}
  \label{fig:system_pipeline}
\end{figure*}

\section{Method: Natural Language MARL}
\label{sec:method}
In this section, we introduce LangMARL, a framework designed to optimize multi-agent LLM systems through credit assignment mechanisms. We first describe the overall paradigm, followed by the specific designs of the language policy actor, the centralized critic, and the language-based credit assignment and optimization, as illustrated in Fig.~\ref{fig:system_pipeline}.
\subsection{Language CTDE Paradigm}
\label{sec:method:paradigms}

We adopt a cooperative multi-agent reinforcement learning setting as introduced in the preliminaries, and instantiate its core components directly in natural language. LangMARL follows a centralized-training–decentralized-execution paradigm, in which credit assignment and policy optimization are implemented in language space.

During training, a centralized language critic has access to the complete episodic trajectory and produces agent-specific credit assignments expressed in natural language. These credits serve as structured, agent-specific training signals that summarize agents’ causal contributions across the entire trajectory. During execution, each agent acts independently using only its language policy, without access to global observations.

\subsection{Language Policy Actor}
\label{sec:method:policy}

Each agent $i$ is instantiated as a large language model equipped with a \emph{language policy}. Formally, we define a text-parameterized policy:
\begin{equation}
    \pi_i^{\text{text}} : \mathcal{S}_i^{\text{text}} \rightarrow \mathcal{A}_i^{\text{text}}
\end{equation}
where $\mathcal{S}_i^{\text{text}}$ denotes a natural-language encoding of agent $i$’s observation, and $\mathcal{A}_i^{\text{text}}$ denotes the space of action descriptions available to agent $i$.

At timestep $t$, agent $i$ samples its action via
$
a_t^i \sim \pi_i^{\text{text}}(\cdot \mid s_t^{\text{text}}),
$
which is implemented in practice by querying the LLM, conditioned on the current policy text and state description:
\begin{equation}
    a_t^i = \text{LLM}_{\text{actor}}\big(\pi_i^{\text{text}}, s_t^{\text{text}}\big)
\end{equation}
From a reinforcement learning perspective, $\pi_i^{\text{text}}$ plays the same role as a stochastic policy $\pi_{\theta_i}$ in classical, parameterized MARL formulations. Its parameters, however, are implicitly encoded in natural language instructions, rules, or exemplars rather than numeric vectors.

\subsection{Centralized Language Critic for Credit Assignment}
\label{sec:method:value}

Consistent with Monte Carlo policy evaluation, LangMARL evaluates policies using complete episodic rollouts without temporal-difference bootstrapping. Given a trajectory
$
\tau = (s_0, \mathbf{a}_0, s_1, \mathbf{a}_1, \dots, s_T),
$
we employ a centralized \emph{language critic}, implemented as an LLM, that consumes the full episodic trajectory and performs causal attribution of individual agent contributions.

Rather than estimating a scalar value function, the language critic performs structured credit assignment by generating agent-specific feedback: 
\begin{equation}
\label{eq:critic}
    C_i^{\text{text}}(\tau) = \text{LLM}_{\text{critic}}\big(\tau, i\big)
\end{equation}
where $C_i^{\text{text}}(\tau)$ describes how agent $i$'s sequence of actions $\{a_t^i\}_{t=0}^{T-1}$ positively or negatively influenced the final outcome of the episode.

\begin{algorithm}[t]
\caption{LangMARL: Centralized Training}
\label{alg:langmarl_train}
\begin{algorithmic}[1]
\REQUIRE
Trajectory set $\{\tau_k\}_{k=1}^K$;  
Language policies $\{\pi_i^{\text{text}}\}_{i=1}^N$;  
Centralized language critic $\text{LLM}_{\text{critic}}$;  
Language policy gradient estimator $\text{LLM}_{\text{grad}}$;  
Language policy optimizer $\text{LLM}_{\text{opt}}$

\STATE \textit{\#Agent-wise Credit Assignment}
\FOR{each trajectory $\tau_k$}
    \FOR{each agent $i \in \{1,\dots,N\}$}
        \STATE Generate language credit with Eq.~\eqref{eq:critic}
    \ENDFOR
\ENDFOR

\STATE \textit{\# Language Policy Gradient Estimation}
\FOR{each agent $i \in \{1,\dots,N\}$}
    \FOR{$k = 1$ to $K$}
        \STATE Compute language gradient with Eq.~\eqref{eq:gradient}
     \ENDFOR
\ENDFOR

\STATE \textit{\#Language Policy Optimization}
\FOR{each agent $i \in \{1,\dots,N\}$}
    \STATE Update language policy with Equation~\eqref{eq:opt}
\ENDFOR

\STATE \textbf{Output:} Updated language policies $\{\pi_i^{\text{text}}\}_{i=1}^N$
\end{algorithmic}
\end{algorithm}


\subsection{Language Policy Gradient Estimator}
\label{sec:method:gradient}

In classical policy gradient methods, policy improvement for each agent is driven by optimizing:

\begin{equation}
\small
    \nabla_{\theta_i} J(\pi_i)
=
\mathbb{E}_{\tau \sim \pi}
\left[
\sum_{t=0}^{T-1}
\nabla_{\theta_i} \log \pi_{\theta_i}(a_t^i \mid s_t)
\cdot G_i(\tau)
\right]
\end{equation}
where $G_i(\tau)$ is a return or advantage estimate.

In LangMARL, actor-side policy updates are driven by a dedicated \emph{policy gradient estimator}, implemented as an LLM. For each trajectory \( \tau_k \) and agent \( i \), we generate a \emph{language policy gradient}:


\begin{equation}
\label{eq:gradient}
\Delta \pi_i^{\text{text}}(\tau_k)
=
\text{LLM}_{\text{grad}}
\Big(
\pi_i^{\text{text}},
C_i^{\text{text}}(\tau_k)
\Big)
\end{equation}
The estimator conditions on the current policy, the agent’s state–action sequence, and the language credit assigned to the trajectory. The output \( \Delta \pi_i^{\text{text}}(\tau_k) \) is a language-form update direction that specifies how the policy should be adjusted in response to the experience \( \tau_k \). This module serves as a language-space analogue of policy gradient estimation, without computing numeric gradients.

\subsection{Language Policy Optimizer}
\label{sec:method:improvement}
To perform batch policy optimization, LangMARL applies a single policy update based on language-form gradient information collected from multiple trajectories. For each agent \( i \), we obtain a set of trajectory-level language gradients
\(
\{\Delta \pi_i^{\text{text}}(\tau_k)\}_{k=1}^K
\),
as described in Section~\ref{sec:method:gradient}. Both gradient aggregation and policy update are implemented using large language models. Specifically, we employ an LLM-based \emph{aggregator},
denoted by \( \text{LLM}_{\text{agg}} \), to semantically integrate multiple trajectory-level gradients into a single update direction, and an LLM-based \emph{language policy optimizer},
denoted by \( \text{LLM}_{\text{opt}} \), to apply the resulting update to the policy.

Formally, the policy update is expressed as a composition of two LLM operators:

\begin{equation}
\small
\label{eq:opt}
    \pi_i^{\text{text}}
\leftarrow
\text{LLM}_{\text{opt}}
\Big(
\pi_i^{\text{text}},\;
\text{LLM}_{\text{agg}}
\big(
\{\Delta \pi_i^{\text{text}}(\tau_k)\}_{k=1}^K
\big)
\Big)
\end{equation}
The aggregation operator \( \text{LLM}_{\text{agg}} \) serves as a language-space analogue of batch gradient averaging. Rather than performing arithmetic averaging, it semantically integrates multiple update directions by identifying consistent modification patterns, resolving conflicts among suggestions, and suppressing noisy or idiosyncratic gradients.

In general, \method organizes policy optimization in natural language into a batch actor–critic style procedure. Trajectory-level update directions are first aggregated and then applied via LLM-based operators, reflecting the functional separation between gradient aggregation and optimizer steps in classical batch policy gradient methods, while operating in a language space, as summarized in Algorithm~\ref{alg:langmarl_train}, and the specific process for trajectory collection is outlined in Appendix~\ref{app:alg_execution}.

%% file: sec_arxiv/5_exp.tex
\begin{table*}[t]
\centering
\caption{Comprehensive Comparison across QA, Reasoning, Coding and Multi-Agent Environments. All reported values represent the mean performance after 5 training iterations using the same backbone LLMs.} 
\label{tab:main_results}
\small
\resizebox{\linewidth}{!}{
\begin{tabular}{l ccc cccc c}
\toprule
\textbf{Category} & \multicolumn{2}{c}{\textbf{Static-Prompting}} & \multicolumn{5}{c}{\textbf{Self-Evolving}} & \multicolumn{1}{c}{\textbf{Ours}} \\ \cmidrule(r){2-3} \cmidrule(r){4-8} \cmidrule(r){9-9}

\textbf{Method} & CoT & Agents & AutoPE & DSPy & Reflexion & TextGrad & Symbolic & LangMARL \\
\hline
\rowcolor[gray]{0.95} \multicolumn{9}{c}{\textit{Natural Language Benchmarks (Accuracy / Pass Rate) }} \\ 
HotPotQA & 38.8 & 37.5 & 39.8 & 43.9 & 59.1 & 57.3 & 44.8 & \textbf{60.2} \\
MATH & 23.2 & 23.8 & 22.5 & 17.3 & 49.4 & 53.8 & 38.8 & \textbf{56.0} \\
HumanEval & 59.2 & 59.5 & 63.5 & 66.7 & 70.1 & 68.9 & 64.5 & \textbf{73.2} \\
\hline
\rowcolor[gray]{0.95} \multicolumn{9}{c}{\textit{Multi-Agent Game: Overcooked-AI (Mean Reward)}} \\
Forced Coord. & 73.3 & 68.4 & 71.9 & 62.1 & 138.5 & 85.3 & 104.2 & \textbf{148.9} \\
Coord. Ring & 140.0 & 122.3 & 133.4 & 141.4 & 157.8 & 163.8 & 148.5 & \textbf{184.4} \\
Counter Circuit & 40.0 & 43.6 & 38.2 & 52.4 & 68.7 & 54.5 & 55.1 & \textbf{77.8} \\
Asymm. Adv. & 226.7 & 202.7 & 143.5 & 217.3 & 204.1 & 230.6 & 232.4 & \textbf{244.4} \\
Cramped Room & 126.7 & 118.5 & 122.8 & 145.7 & 151.1 & 131.3 & 147.5 & \textbf{171.4} \\
\hline
\rowcolor[gray]{0.95} \multicolumn{9}{c}{\textit{Multi-Agent Game: Pistonball (Mean Reward)}} \\
$N=10$ & -0.9 & 11.6 & 5.8 & 20.0 & 23.7 & 33.5 & 21.1 & \textbf{37.2} \\
$N=12$ & -0.1 & -2.7 & -0.8 & 7.8 & -6.2 & 24.3 & 3.1 & \textbf{29.6} \\
$N=16$ & -8.5 & -1.0 & -3.1 & -0.9 & 8.9 & 10.1 & -7.6 & \textbf{17.2} \\
$N=20$ & -11.5 & -3.8 & -6.2 & -7.3 & -1.2 & 14.3 & 3.3 & \textbf{22.9} \\
\bottomrule
\end{tabular}
}
\end{table*}

\section{Experiments}
We conduct comprehensive experiments across diverse domains including code generation, multi-hop reasoning, mathematics, and strategic games to evaluate the efficacy of LangMARL. Our  evaluation is guided by the following Research Questions:

\noindent$\bullet$~\textbf{RQ1: Overall Effectiveness.} Does LangMARL enhance team performance in cooperative tasks compared to other baselines?
    
\noindent$\bullet$~\textbf{RQ2: Credit Assignment Quality.} How effectively does the Centralized Language Critic disentangle individual agent contributions compared to global reflection?
    
\noindent$\bullet$~\textbf{RQ3: Robustness and Scalability.} How does LangMARL's performance scale with respect to: (i) the reasoning capacity of the underlying LLM, and (ii) the volume of trajectory data available for policy evolution?

\begin{figure*}[htbp]                                
      \centering            
      \includegraphics[width=\linewidth]{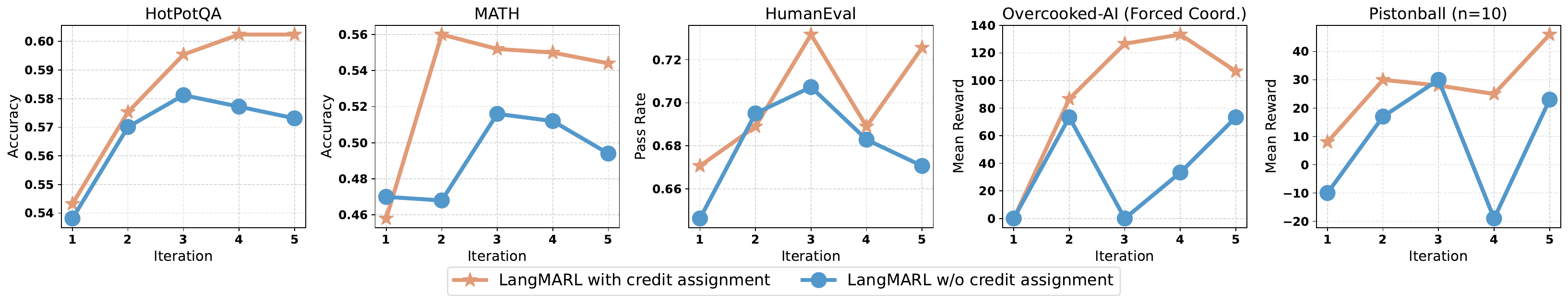}           
      \vspace{-8mm}
      \caption{\textbf{Impact of credit assignment on the convergence quality of LangMARL.} The learning curves across five benchmark tasks demonstrate that credit assignment is pivotal for efficient policy optimization and superior final performance. Without this mechanism (green lines), the models exhibit sub-optimal learning rates and significant instability, particularly in complex reasoning and multi-agent coordination scenarios.}
      \vspace{-5mm}
      \label{fig:training_curve}
  \end{figure*}

\begin{figure*}[t!]
  \centering
  \includegraphics[width=1.0\textwidth]{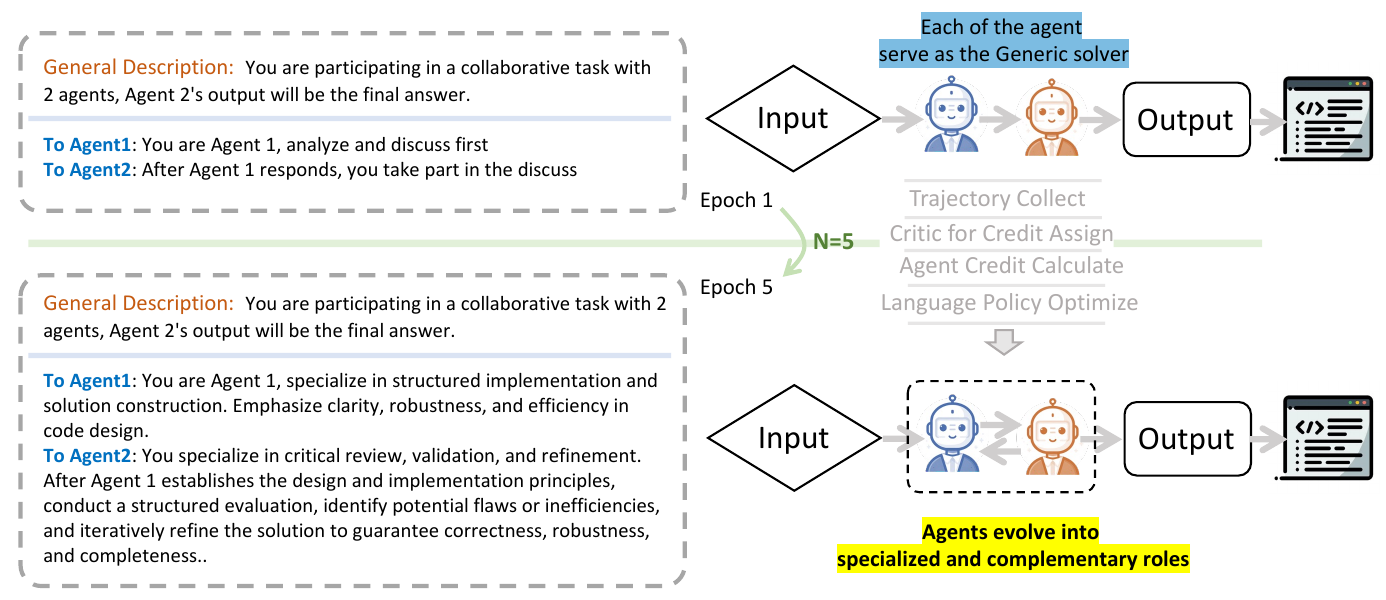}
  \caption{
\textbf{Emergent role specialization in LangMARL.} 
Top: In the initial symmetric setup, both agents operate as generic problem solvers without predefined role differentiation. 
Bottom: After iterative trajectory-level credit assignment and language-based policy optimization, agents self-organize into complementary roles, such as in the coding scenario, Agent~1 specializing in structured implementation and Agent~2 in critical evaluation and refinement. This division of labor is not explicitly specified in the prompt but emerges through centralized language credits. 
}
  \label{fig:pattern_nlp}
\end{figure*}

\subsection{Environment Settings}

We evaluate LangMARL across two distinct categories of environments (see Appendix~\ref{app:settings}). 

\noindent$\bullet$~\textbf{Strategic Game Environments:}
Strategic games feature fixed agent roles and physical constraints, providing a controlled setting to evaluate credit assignment ability. 
\textit{Overcooked-AI} \cite{carroll2019utility} is a benchmark for coordination where two agents collaborate to cook and deliver soup. We employ a hierarchical action space where high-level semantic subgoals are translated into primitive movements. \textit{Pistonball} \cite{terry2021pettingzoo} involves a team of  pistons working together to move a ball. Each agent has severe partial observability, seeing only a local vertical slice of the environment.

\noindent$\bullet$~\textbf{Cooperative Language Tasks:}
To assess LangMARL’s performance in open-task environments, we reframe three prominent NLP datasets \textit{HumanEval} (coding), \textit{HotPotQA} (multi-hop reasoning), and \textit{MATH} (mathematics) into a multi-agent reinforcement learning framework. Unlike the fixed-role games, these tasks evaluate the framework's capability to foster the emergence of novel coordination strategies. 

\subsection{Baselines and Comparisons}

We compare LangMARL with prompting and self-evolving methods listed in Table~\ref{tab:self-evolving-compare}. 
Specifically, these baselines can be categorized in two groups:
(1) \textit{Static prompting.}
CoT~\cite{wei2022chain} and Agents~\cite{zhou2023agents} use fixed prompts without adaptive updates.
(2) \textit{Prompt optimization.}
AutoPE~\cite{zhou2023large} and DSPy~\cite{khattab2023dspy} iteratively refine prompts, but optimize each agent independently. Reflexion~\cite{shinn2023reflexion} and TextGrad~\cite{yuksekgonul2024textgrad} update prompts using linguistic feedback. Symbolic Learning~\cite{ou2025symbolic} enables multi-agent optimization and relies on team-level signals.

\begin{figure*}[t!]
  \centering
  \includegraphics[width=1.0\textwidth]{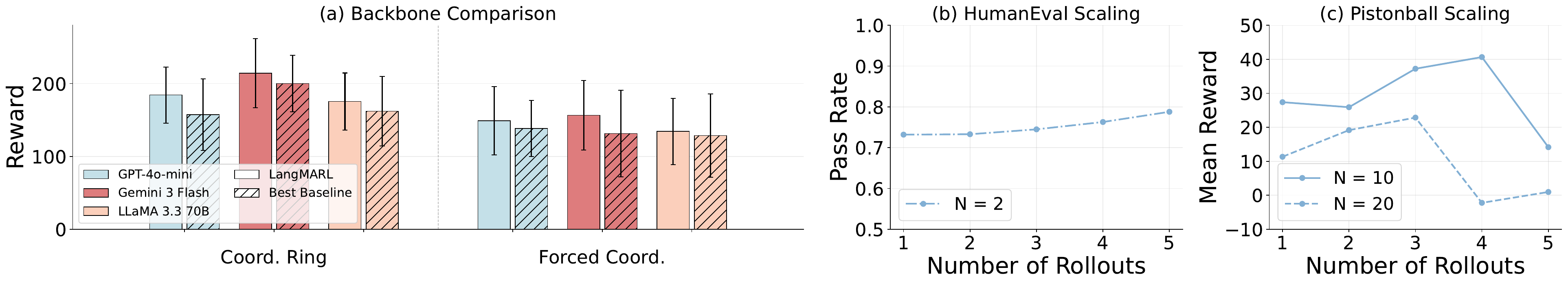}
  \vspace{-8mm}
  \caption{\textbf{Sensitivity analysis of LangMARL components.} (a) Performance comparison of various LLM backbones against the best baseline Reflexion. (b-c) Impact of Monte Carlo rollout counts on HumanEval and Pistonball.} 
  \label{fig:analysis_experiments}
  \vspace{-3mm}
\end{figure*}

\subsection{Results of Overall Performance (RQ1)}
In this experiment, following existing work~\cite{ou2025symbolic, khattab2023dspy, shinn2023reflexion}, GPT-3.5-turbo is employed for natural language benchmarks, while GPT-4o-mini is utilized for both game environments. As summarized in Table~\ref{tab:main_results}, \method achieves superior performance across all benchmarks, consistently outperforming both static prompting and self-evolving baselines. 

\paragraph{Natural Language Benchmarks.} 
On reasoning-intensive tasks, \method significantly surpasses other optimization methods on \textit{MATH}, \textit{HotPotQA} and \textit{HumanEval} datasets. These results demonstrate that addressing the credit assignment problem during interaction optimization allows for robust performance gains across diverse benchmarks. Notably, the performance on these benchmarks exhibits a strong positive correlation with rollout budgets, suggesting that \method effectively translates increased computational resources into reasoning improvements.

\paragraph{Strategic Game Environments.} 
In \textit{Overcooked-AI}, \method attains the highest mean reward across all layouts. In \textit{Pistonball}, \method maintains stable positive returns as team size scales to $N=20$, whereas baselines like TextGrad degrade. This confirms that our centralized critic effectively enables scalable cooperation under partial observability and sparse rewards. Furthermore, \method demonstrates superior sample efficiency in large-scale coordination tasks. 

\subsection{Credit Assignment Quality (RQ2)}
We validate the efficacy of the Language Critic through learning stability and qualitative emergent behaviors.
\begin{wrapfigure}{r}{0.5\textwidth}
  \centering
  \includegraphics[width=0.48\textwidth]{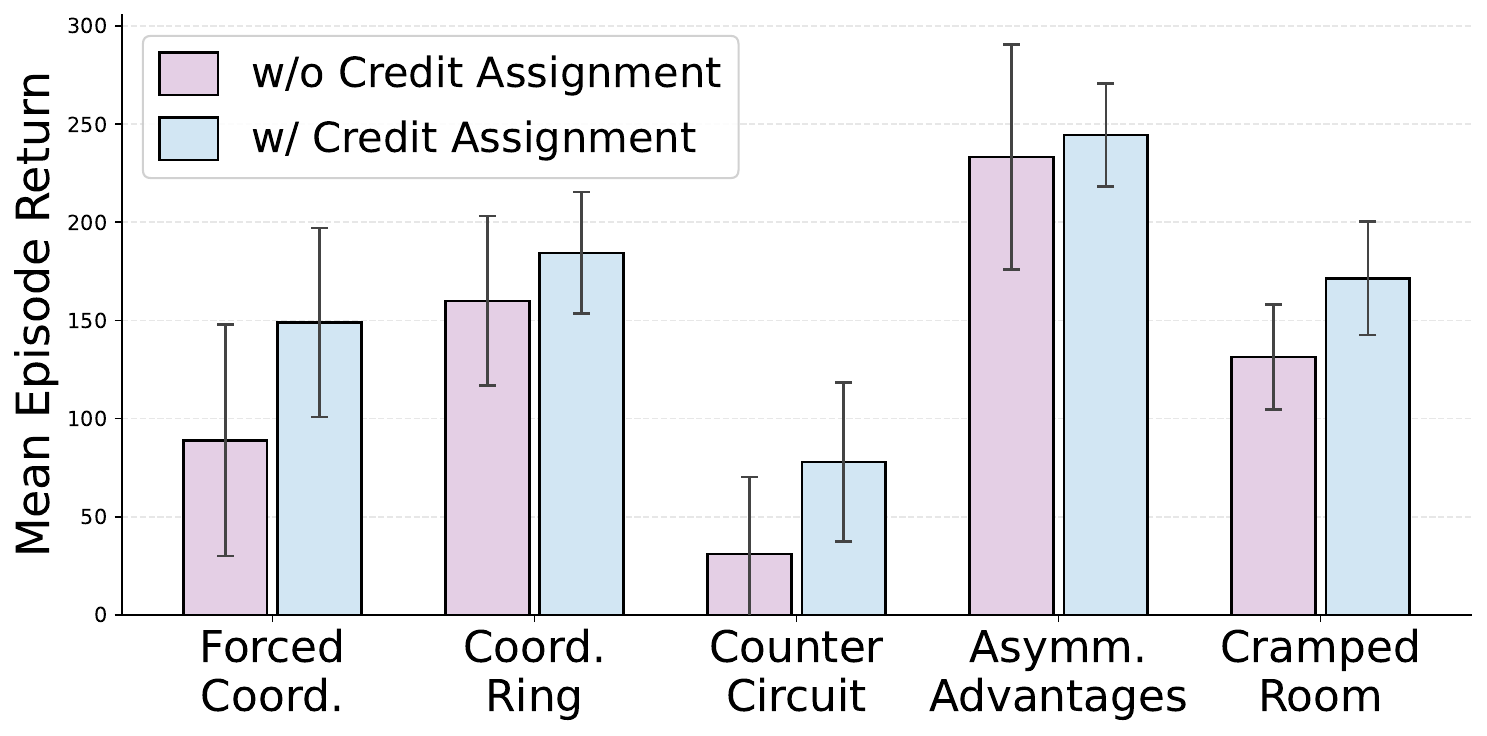}
  \caption{Ablation study on credit assignment.}
  \vspace{-15mm}
  \label{fig:ablation}
\end{wrapfigure}

\vspace{-5mm}
\paragraph{Ablation on Credit Assignment.} Fig.~\ref{fig:ablation} evaluates the impact of explicit agent-wise credit assignment. Removing credit assignment and training all agents with a shared global reward leads to a clear performance drop in both Overcooked-AI and Pistonball. These results confirm that explicit credit assignment is a key component for stable learning and effective coordination in LangMARL.

\paragraph{Learning Dynamics and Stability.} 
As shown in Fig.~\ref{fig:training_curve}, centralized credit assignment (blue lines) is pivotal for stable optimization. Across all benchmarks, its inclusion leads to faster convergence and higher final performance, whereas models without it (green lines) suffer from learning instability and suboptimal plateaus.

\paragraph{Emergent Role Specialization.} 
We showcase that \method can automatically induce functional specialization without predefined roles:
(1)~\textbf{Collaborative Coding:} Fig.~\ref{fig:pattern_nlp} shows Agents evolve from symmetric solvers into complementary roles over five epochs. Agent 1 focuses on \textit{implementation} while Agent 2 specializes in \textit{critical review and refinement}.
(2)~\textbf{Strategic Games:} Fig.~\ref{fig:pattern_game} illustrates the critic resolving bottlenecks. In \textit{Overcooked-AI}, Player Green optimizes resource placement based on identified needs of Player Blue. In \textit{Pistonball}, the critic identifies and corrects geometric blockages caused by specific pistons.

\subsection{Sensitivity Analysis (RQ3)}
We evaluate robustness across model backbones and rollout budgets.

\paragraph{Backbone Robustness.} 
As illustrated in Fig.~\ref{fig:analysis_experiments}(a), \method demonstrates strong compatibility across various LLM backbones. In both layouts of \textit{Coord. Ring} and \textit{Forced Coord.}, \method consistently achieves higher mean rewards compared to the strongest baselines, regardless of the underlying model. Notably, Gemini-3-Flash paired with our framework delivers the peak performance, while even the open-weight LLaMA-3.3-70B architecture maintains competitive results, underscoring \method's robustness and its generalizability across different scales and families of LMs.

\paragraph{Impact of Rollout Budget.} 

Furthermore, the impact of rollout budgets varies significantly with task complexity and agent scale. In few-agent tasks such as \textit{HumanEval} (Fig.~\ref{fig:analysis_experiments}b), increasing rollouts leads to a consistent and significant performance gain. However, in the large-scale multi-agent \textit{Pistonball} environment (Fig.~\ref{fig:analysis_experiments}c), the benefit of an increased budget is highly sensitive to the team size $N$. While performance for a large team ($N=10$ or $20$) initially improves, it subsequently reaches a plateau. This suggests an optimal rollout frequency exists, as excessive rollouts in large-scale systems can exacerbate policy bias despite improving coverage in simpler settings. 

%% file: sec_arxiv/6_conclusion.tex
\section{Conclusion}
This paper establishes a framework designed to optimize multi-LLM agents through credit assignment mechanisms inspired by MARL. Our experiments across strategic games and reasoning benchmarks demonstrate that \method consistently outperforms other self-evolving baselines. Ablation studies and learning dynamics confirm that agent-specific credit assignment is the primary driver of optimization stability, enabling the spontaneous emergence of functional specialization without manual engineering. Furthermore, sensitivity analysis shows that these performance gains are robust across diverse LLM backbones. Future research will explore hierarchical sub-goal decomposition and dynamic agent synthesis to facilitate more fluid, large-scale autonomous collaboration. Despite the promising results, several challenges remain to be addressed:




\paragraph{Scaling to Long-horizon Agentic Tasks.}
A primary limitation is \method's performance in long-horizon workflows, such as autonomous software development, where sparse rewards and error propagation over hundreds of steps pose significant challenges. To address this, future work could investigate the hierarchical decomposition of extended trajectories into manageable semantic sub-goals. This approach would allow the centralized critic to provide more granular feedback, maintaining the consistency of semantic credit assignment even when the causal link between an initial intervention and the final outcome is significantly attenuated over time.

\paragraph{Dynamic Sub-agent Synthesis and Orchestration.}
Currently, our framework operates within a pre-defined agent topology, which may limit its adaptability in real-world collaboration requiring fluid organizational structures. We aim to extend \method toward "on-the-fly" synthesis and instantiation of new agents. A potential solution involves developing mechanisms for dynamic task partitioning and the evolution of hierarchical communication protocols. This would enable the system to autonomously spawn specialized sub-agents and adaptively scale its computational resources in response to emerging sub-tasks, a direction we plan to explore in subsequent studies.

\section{Acknowledgement}
The work was partially supported by Cisco Faculty Award, NSF award \#2442477 and \#2550203. The views and conclusions in this paper should not be interpreted as representing any funding agencies.

%% file: sec_arxiv/appendix.tex
\clearpage

\appendix

\section{Appendix}

\begin{figure*}[t!]
      \centering
  \includegraphics[width=1.0\textwidth]{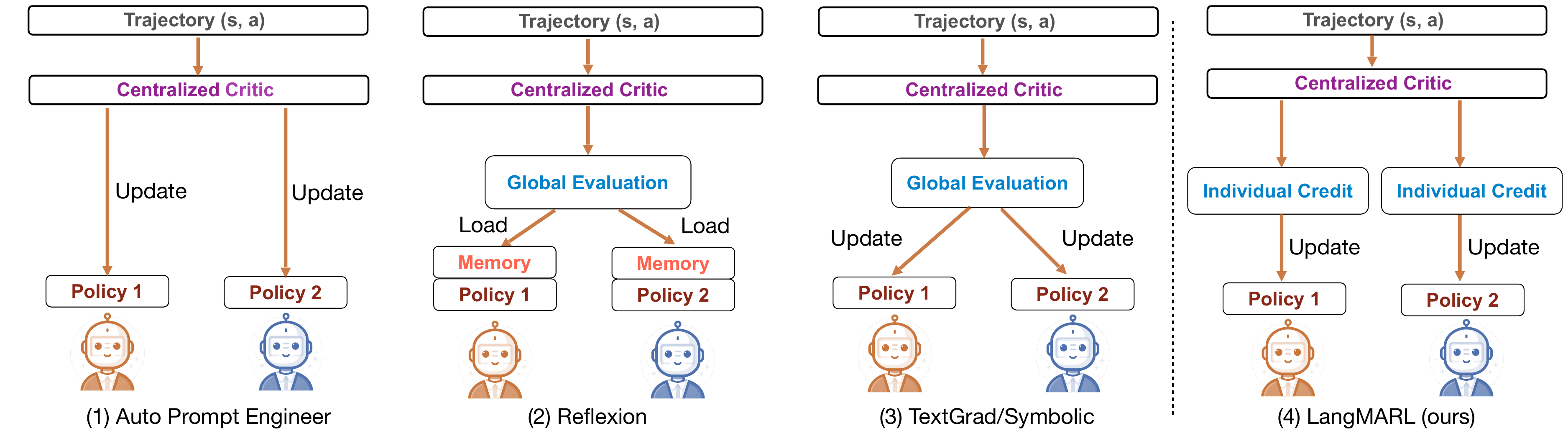}
  \caption{Comparison of optimization paradigms in multi-agent systems. While existing methods rely on direct updates (1), shared memory (2), or collective global evaluations (3), LangMARL (4) introduces an explicit individual credit assignment mechanism. This allows the centralized critic to decompose global feedback into agent-specific guidance, facilitating more precise policy optimization in collaborative tasks.} 
  \label{fig:paradigm}
\end{figure*}


\subsection{Paradigm Comparison}
\label{app:paradigm}

We compare \method against existing optimization paradigms (Fig.~\ref{fig:paradigm}) to highlight the evolution toward precise multi-agent coordination. Early methods like \textbf{Auto Prompt Engineer (1)} treat policies as black boxes, relying on global outcomes to propose updates; however, they lack the internal feedback loops necessary for complex joint strategy spaces. \textbf{Reflexion (2)} introduces a shared memory for agents to process past failures, yet these reflections are often monolithic, leading to a "bystander effect" where individual agents cannot distinguish their specific contributions to a collective error. \textbf{TextGrad and Symbolic (3)} advance this by utilizing LLMs for global evaluations, but they typically generate a single, joint feedback signal. In tasks like Overcooked or Pistonball, this prevents the system from pinpointing specific bottlenecks, often causing "credit drift" where a high-performing agent's policy is incorrectly modified due to a teammate's mistake. In contrast, \textbf{\method (4)} introduces an explicit individual credit assignment mechanism. By utilizing a Centralized Language Critic to decompose global trajectories into agent-specific "Language Gradients," our approach captures the causal dependencies between agents (e.g., identifying when one agent's action blocks another). This granular feedback transforms global rewards into localized, actionable guidance, preventing coordination collapse and enabling the spontaneous emergence of the functional specializations observed in our experiments.

\begin{figure*}[t]                                 
      \centering      
      \includegraphics[width=\linewidth]{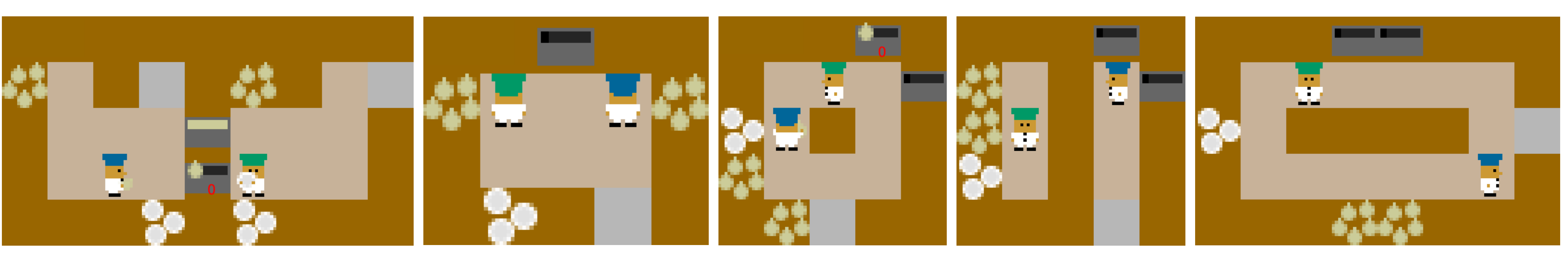}  
      \caption{\textbf{Overcooked-AI Layouts.} From left to right: Asymmetric Advantages tests the adoption of high-level strategies based on individual strengths; Cramped Room presents low-level coordination challenges within a confined, collision-prone space; Coordination Ring requires synchronized navigation between opposite corners; Forced Coordination removes collision friction to focus on joint strategy, as neither agent can serve dishes alone; and Counter Circuit necessitates a non-obvious strategy of passing ingredients over counters rather than carrying them.}
      \label{fig:overcooked_layouts}
  \end{figure*}

\begin{figure*}[t]                                 
      \centering      
      \includegraphics[width=\linewidth]{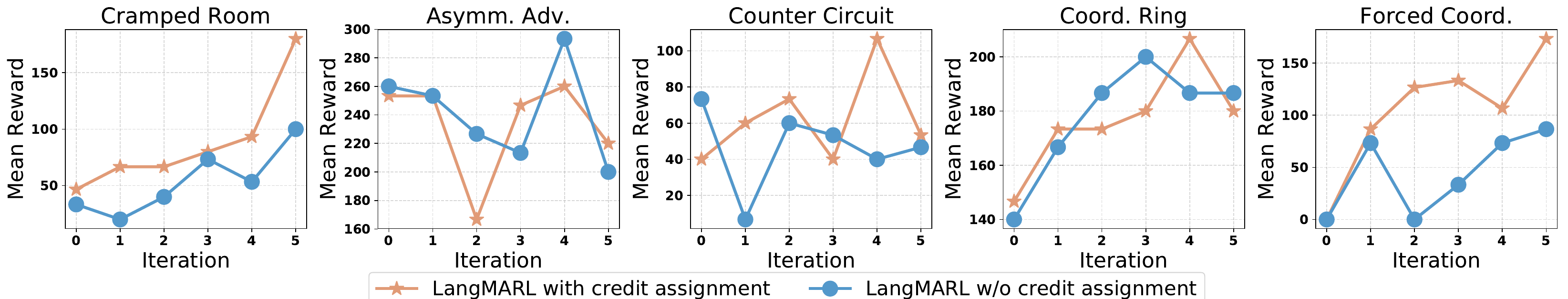}  
      \caption{Learning curves on Overcooked-AI layouts. Comparison of LangMARL with (orange) and without (blue) credit assignment. Results across five maps show that language-level credit assignment significantly improves mean reward and convergence stability, especially in layouts requiring high-degree coordination (e.g., Coord. Ring and Forced Coord.).}
      \label{fig:results}
  \end{figure*}

\subsection{Detailed Implementation of Experience Collection}
\label{app:alg_execution}

In this section, we present the formal procedure for the decentralized experience collection phase of LangMARL, as summarized in Algorithm \ref{alg:langmarl_collect}. This stage is crucial for operationalizing the Decentralized Execution part of the CTDE paradigm within a language space.Process Overview: During each training iteration, the system performs $K$ independent rollouts. Each agent $i$ maintains its own decentralized language-based policy $\pi_i^{\text{text}}$, which consists of a set of textual instructions and few-shot examples. At each time step $t$, agents process the current textual state $s_t^{\text{text}}$—which includes environmental observations and potential communication history—to sample a discrete action $a_t^{i,k}$ in parallel.

\begin{algorithm}[htbp]
\caption{LangMARL: Decentralized Experience Collection} 
\label{alg:langmarl_collect}
\begin{algorithmic}[1]
\REQUIRE

Number of agents $N$;  
Language policies $\{\pi_i^{\text{text}}\}_{i=1}^N$;  
Number of rollouts $K$;  
Episode horizon $T$

\FOR{$k = 1$ to $K$}
    \STATE Initialize environment
    \STATE Obtain initial textual state $s_0^{\text{text}}$
    \FOR{$t = 0$ to $T-1$}
        \FOR{each agent $i \in \{1,\dots,N\}$ \textbf{in parallel}}
            \STATE Sample action from language policy: 
            $ a_t^{i,k} \sim \pi_i^{\text{text}}(\cdot \mid s_t^{\text{text}}) $
        \ENDFOR
        \STATE Execute joint action $\mathbf{a}_t^k = (a_t^{1,k},\dots,a_t^{N,k})$
        \STATE Observe next textual state $s_{t+1}^{\text{text}}$
    \ENDFOR
    \STATE Store complete trajectory: 
    $    \tau_k = (s_0^{\text{text}}, \mathbf{a}_0^k, \dots, s_T^{\text{text}})    $
\ENDFOR
\STATE \textbf{Output:} Trajectory set $\{\tau_k\}_{k=1}^K$

\end{algorithmic}
\end{algorithm}

\subsection{Environment Settings}
\label{app:settings}


We briefly summarize the key environment configurations used in our experiments, including the number of agents, observation spaces, action spaces, and reward semantics. 

\noindent$\bullet$~\textbf{Overcooked-AI}~\cite{carroll2019utility} is a cooperative environment in which two agents collaborate on a cooking task, as shown in Fig.~\ref{fig:overcooked_layouts}. Each episode proceeds in synchronized time steps, where both agents act simultaneously. At each step, each agent receives a partial observation consisting of a symbolic, egocentric grid representation of the kitchen layout, including nearby terrain, objects, and the other agent’s position and orientation.

For action modeling, we adopt the hierarchical action space. At the high level, an agent selects a semantic subgoal (e.g., fetch ingredient, deliver soup), while at the low level, the policy outputs primitive actions from a discrete set including movement, interaction, and orientation changes. This two-layer design enables structured temporal abstraction while retaining compatibility with the original Overcooked-AI dynamics. The reward function in Overcooked-AI is sparse and team-based: a positive reward is issued only when a soup is successfully delivered.

\begin{wrapfigure}{r}{0.5\textwidth}               
      \centering      
      \includegraphics[width=\linewidth]{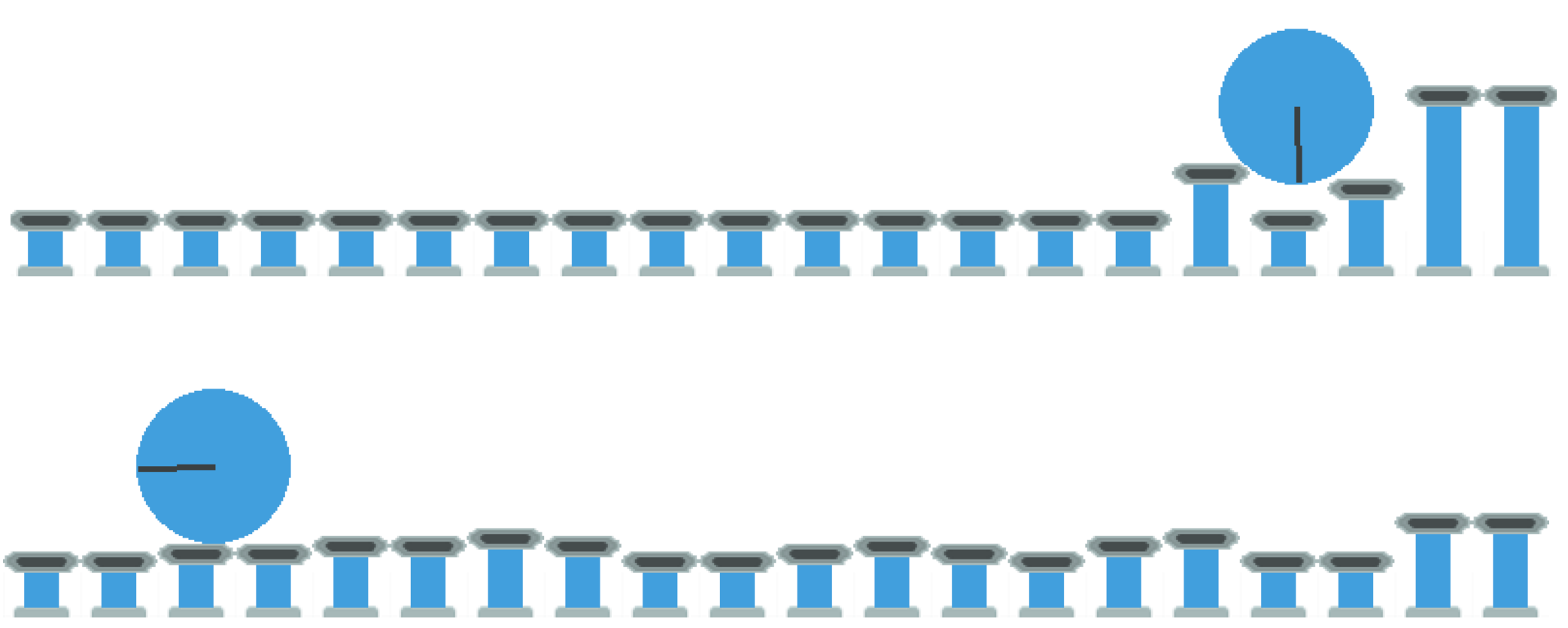}  
      \caption{\textbf{Pistonball Layouts.} The upper panel illustrates a failure case where poor coordination among piston agents leads to an uneven surface, trapping the ball and resulting in a significantly low negative score due to the accumulated time penalty. In contrast, the lower panel demonstrates successful coordination, where agents synchronously adjust their vertical positions to smoothly propel the ball toward the target.}
      \vspace{-10mm}
      \label{fig:pistonball_layouts}
\end{wrapfigure}

\noindent$\bullet$~\textbf{Pistonball}~\cite{terry2021pettingzoo} is a cooperative multi-agent environment with variable team sizes. In our experiments, the number of piston agents ranges from 10 to 20, covering different levels of coordination difficulty. The environment evolves in discrete time steps governed by a physics simulator, as shown in Fig.~\ref{fig:pistonball_layouts}.

Each piston agent receives a local, high-dimensional pixel-based observation capturing a narrow vertical slice of the environment centered around the piston, including the ball's relative position and nearby physical context. No agent has access to global state information, resulting in severe partial observability. At each step, each agent selects from a small discrete set of vertical movement actions, which control the piston's vertical force. All agents share a single global cooperative reward composed of two terms: a distance-based component proportional to the ball's leftward progress in the current time step, and a constant time penalty (default: $-0.1$) applied at every step to incentivize faster task completion. This reward structure encourages collective physical coordination while still requiring agents to infer their individual contributions.

\begin{figure*}[t!]
  \centering
  \includegraphics[width=1.0\textwidth]{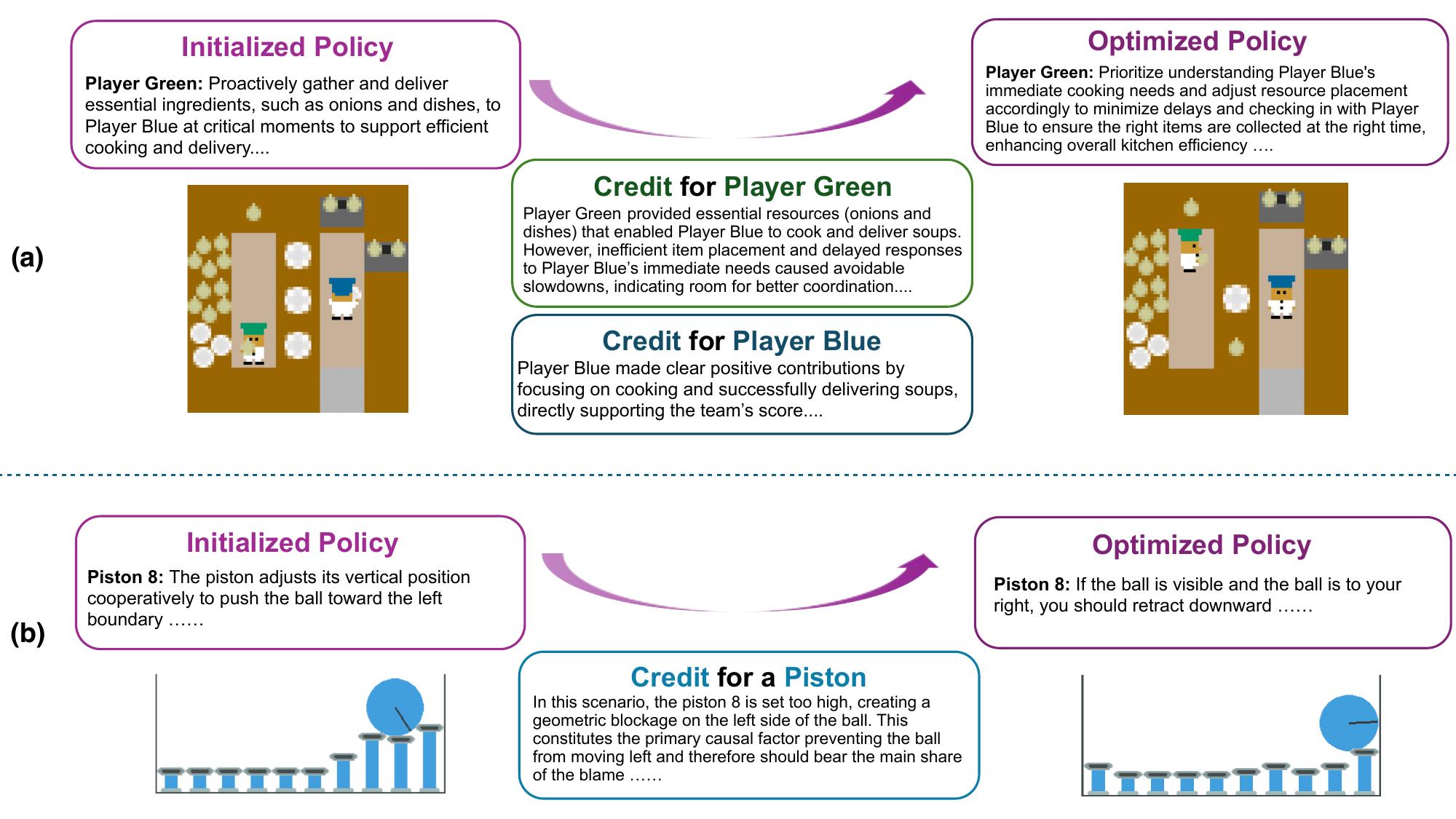}
  \caption{\textbf{Examples of language-based credit assignment and policy optimization in LangMARL.} (a)~Overcooked: the centralized language critic analyzes a full trajectory and produces agent-specific credits, which are then used to revise the language policy of Player Green, resulting in improved coordination with Player Blue. (b)~Pistonball: the language critic identifies a causally responsible piston from the trajectory and assigns targeted credit, leading to a refined local policy that corrects blocking behavior.} 
  \label{fig:pattern_game}
\end{figure*}

\vspace{-5mm}
\subsection{Learning Dynamics across Overcooked Scenarios}
\label{app:learning_curves}

To supplement the primary results, Figure~\ref{fig:results} illustrates the learning curves across four Overcooked-AI layouts. The dynamics vary significantly due to the distinct coordination challenges in each map:

\noindent$\bullet$~\textbf{Cramped Room:} While simpler, it often leads to spatial conflict. LangMARL quickly resolves these "collision" bottlenecks through precise language feedback.
\noindent$\bullet$~\textbf{Asymmetric Advantages:} This scenario necessitates labor division. LangMARL demonstrates a clear "role discovery" capability, where the critic assigns credit based on individual task efficiency, leading to specialized policies.
\noindent$\bullet$~\textbf{Coordination Ring:} LangMARL’s dense, trajectory-based credit assignment enables agents to master material transfer via counters, maintaining stable coordination.
\noindent$\bullet$~\textbf{Forced Coordination:} This is the most difficult layout due to the required item-passing. LangMARL successfully bridges the coordination gap, showing a robust and steady increase in reward.

\subsection{Prompt Templates}
\label{app:prompts}

In this section, we provide the specific prompt templates used for the four core language operators in \method. These templates define the functional roles of language modules in the system and illustrate how policy execution, trajectory evaluation, credit assignment, and policy optimization are realized through natural language interactions.

\begin{tcolorbox}[colback=gray!5,colframe=gray!60!black,title=Prompt Template 1: Language Policy Actor]
\small
\textbf{System Role:} You are a specialized agent [Agent ID] in a cooperative environment. \\
\textbf{Inputs:} 

\noindent$\bullet$~\textbf{Language Policy:} [Current text-based strategy, e.g., ``Focus on supplying ingredients to the chef.'']
\noindent$\bullet$~\textbf{Local Observation:} [Current textual description of the state, e.g., ``You see a tomato at (3,4).'']

\textbf{Instruction:} Based on your current policy and the observation, choose the best action to maximize team reward. \\
\textbf{Output Format:} 

\noindent$\bullet$~\textit{Thinking:} [Reasoning process via Chain-of-Thought]
\noindent$\bullet$~\textit{Action:} [The primitive or semantic action to execute]
\end{tcolorbox}

\begin{tcolorbox}[colback=gray!5,colframe=gray!60!black,title=Prompt Template 2: Centralized Language Critic]
\small
\textbf{System Role:} You are an expert analyst observing a multi-agent team trajectory. \\
\textbf{Inputs:} 

\noindent$\bullet$~\textbf{Joint Trajectory $\tau$:} [Full sequence of states and joint actions of all agents.]
\noindent$\bullet$~\textbf{Team Reward:} [Sparse/delayed reward signal, e.g., ``Task Success'']

\textbf{Instruction:} Perform causal reasoning to disentangle individual contributions. Identify which specific actions by which agent led to the final outcome. \\
\textbf{Output Format:} 

\noindent$\bullet$~\textit{Credit Assignment [Agent i]:} [Detailed linguistic critique of Agent i's causal contribution.]

\end{tcolorbox}

\begin{tcolorbox}[colback=gray!5,colframe=gray!60!black,title=Prompt Template 3: Language Policy Gradient Estimator]
\small
\textbf{System Role:} You are a policy optimization specialist. \\
\textbf{Inputs:} 

\noindent$\bullet$~\textbf{Language Credits:} [Critiques from the Centralized Critic for the specific agent.]
\noindent$\bullet$~\textbf{Prior Policy:} [The agent's current text-based reasoning protocol.]

\textbf{Instruction:} Translate the credit signals into a ``language gradient.'' Specify exactly how the policy should be modified to reinforce positive behaviors or correct failures. \\
\textbf{Output Format:} 

\noindent$\bullet$~\textit{Language Gradient:} [Directional instructions for policy improvement.]

\end{tcolorbox}

\begin{tcolorbox}[colback=gray!5,colframe=gray!60!black,title=Prompt Template 4: Language Policy Optimizer]
\small
\textbf{System Role:} You are a high-level strategy optimizer. \\
\textbf{Inputs:} 

\noindent$\bullet$~\textbf{Aggregated Gradients:} [A set of language gradients derived from multiple trajectories $\{\tau_k\}$.]
\noindent$\bullet$~\textbf{Prior Policy:} [The existing policy to be updated.]

\textbf{Instruction:} Synthesize the gradients across different scenarios. Identify consistent patterns, resolve conflicting feedback, and generate a refined, more robust version of the policy. \\
\textbf{Output Format:} 

\noindent$\bullet$~\textit{Updated Policy:} [The new comprehensive textual strategy for the agent.]

\end{tcolorbox}

\subsection{Qualitative Demonstrations and Case Studies}
\label{app:demos}

To provide a deeper understanding of how LangMARL facilitates coordination, we present qualitative case studies from Overcooked and Pistonball in Figure~\ref{fig:pattern_game}. These examples illustrate the transition from raw trajectory data to actionable language-based policy updates. All agents are instantiated with the same base LLM but guided by different functional instructions to achieve the Centralized Training, Decentralized Execution (CTDE) paradigm in language space, where agents act based on local observations while a centralized critic analyzes global trajectories to generate learning signals for policy evolution.

\textbf{Overcooked.} As shown in Fig.~\ref{fig:challenge} and Fig.~\ref{fig:pattern_game}(a), the agents initially fail because Player Green prioritizes placing plates rather than supplying the onion needed to complete the soup, leaving Player Blue unable to finish cooking. A global reward signal only indicates "order failed," causing both agents to question their own behavior without identifying the true bottleneck (Fig.~\ref{fig:challenge}). In contrast, our Centralized Language Critic identifies the causal link: Player Green's misaligned resource prioritization is responsible for the delay. The resulting per-agent language credit specifically instructs Player Green to "prioritize understanding Player Blue's immediate cooking needs and adjust resource placement accordingly." As shown in the optimized policy (Fig.~\ref{fig:pattern_game}(a)), this targeted credit drives Player Green to proactively coordinate item delivery timing, resolving the bottleneck and improving overall kitchen efficiency.

\textbf{Pistonball.} Correcting Misaligned Actions. In the Pistonball task (Figure~\ref{fig:pattern_game}(b)), coordination often collapses when specific pistons remain elevated, blocking the ball's progress. The critic analyzes the trajectory to identify the causally responsible agent (the specific piston) and generates a "Language Gradient" highlighting the negative impact of its timing. The policy is subsequently optimized to move down earlier, ensuring the ball's momentum is maintained.

These cases demonstrate that semantic credit assignment effectively improves policy optimization. By replacing trial-and-error with causal reasoning in natural language, LangMARL ensures that each agent understands its unique role and contribution within the team, which is the key to scaling to complex multi-agent environments.

\subsection{AI usage}

During the preparation of this work, the authors employed LLMs to assist with code implementation and to refine the manuscript's grammatical clarity and stylistic flow. All final content, technical contributions, and experimental analyses were rigorously reviewed and verified by the authors.